# Failure-Free Genetic Algorithm Optimization of a System Controller Using SAFE/LEARNING Controllers in Tandem

E.S. Sazonov, D. Del Gobbo, P. Klinkhachorn and R. L. Klein
Lane Dept. of Computer Science and Electrical Engineering, West Virginia University,
Morgantown, WV 26506-6109

*Abstract* — **The paper presents a method for failure-free genetic algorithm optimization of a system controller. Genetic algorithms present a powerful tool that facilitates producing near-optimal system controllers. Applied to such methods of computational intelligence as neural networks or fuzzy logic, these methods are capable of combining the non-linear mapping capabilities of the latter with learning the system's behavior directly, that is, without a prior model. At the same time, genetic algorithms routinely produce solutions that lead to the failure of the controlled system. Such solutions are generally unacceptable for applications where safe operation must be guaranteed. We present here a method of design, which allows failure-free application of genetic algorithms through utilization of SAFE and LEARNING controllers in tandem, where the SAFE controller recovers the system from dangerous states while the LEARNING controller learns its behavior. The method has been validated by applying it to an inherently unstable system – inverted pendulum.**

## I. INRODUCTION

The traditional approach to building system controllers requires a prior model of the system. The quality of the model, that is, loss of precision from linearization and/or uncertainties in the system's parameters negatively influence the quality of the resulting control.

At the same time, methods of soft computing such as neural networks or fuzzy logic possess non-linear mapping capabilities, do not require an analytical model and can deal with uncertainties in the system's parameters. Combined with the evolutionary learning (such as genetic algorithms) these methods are capable of producing near-optimal controllers for the given control task. For example, genetic algorithms have been used to produce parameters of an optimized system controller such as the architecture and/or weights of neural network controller [1,2], rules and/or membership functions of a fuzzy controller [3,4], model equations [5], etc.

The disadvantage of the Genetic Algorithms (GA) is that the process routinely produces solutions (parameter sets of a controller) that may render the controlled system unstable. This fact limits applicability of the genetic algorithms to systems where a failure is unacceptable or has a very high associated cost.

This paper shows how evolutionary learning can be applied to produce an optimized neural controller for an open-loop unstable system. The suggested method ensures a failure-free optimization process.

## II. TEST BED

A numerical model of an Inverted Pendulum (IP) served as the test bed for the development of the proposed methodology. Utilization of a model instead of an actual system allowed expediting and simplifying the experimentation process. The performance and accuracy of the model was verified during the design of a classical controller [6].

The IP system consists of a cart sliding on a rail and a rod pivoted to the cart and free to rotate about an axis perpendicular to the direction of motion of the cart. The system is equipped with two sensors measuring cart position and rod angle, and a DC motor providing motion control. The numerical model of the IP system not only simulates the dynamics of IP motion, including saturations on the state variables of cart position and rod angle, but also accounts for major non-linearities of the system, including the dead zone and saturation of the DC motor input voltage and force it can produce. Additionally, the model incorporates such parameters as sensor offsets, discretization errors and measurement noise. More details, including corresponding modeling equations can found in [6].

## III. EXPERIMENTAL SETUP

This paper describes a method for failure-free genetic algorithm optimization of a LEARNING controller, which is achieved through utilization of a SAFE controller. The practical task was to produce a neural controller that has, if possible, a better steady state performance (expressed in the terms of RMS error of the cart position and rod angle during rod balancing) than the original SAFE controller and to prevent the IP system from failing during the experiments.

The SAFE controller is a controller that has been validated for performance and stability of operation, though it might not be an optimal controller. The SAFE controller provides a control design, which is ensured to be failure-free even in cases in which the GA optimization process may generate unacceptable solutions.

The LEARNING controller is a controller being optimized by the genetic algorithms. At different times the LEARNING controller has parameters specified by different specimens of the GA population. Usually, a specimen (genetic "DNA") will define characteristics of the LEARNING controller for a time period *T*, sufficient to determine the fitness value of that specimen. After the period *T*, the next specimen will be used to set parameters of the LEARNING controller and so on. After all specimens in the population have been tested, the population undergoes genetic manipulations, such as reproduction, crossover and mutation [7]. All these manipulations are random in nature and the results of such manipulation can only be determined by testing the new generation on the controlled system. Although the average (on a specimen population) quality of control will improve in time, the nature of the genetic optimization will create faulty LEARNING controllers at each generation. Therefore, despite continuous improvement in quality of control obtained by the best-fitted specimen in the population, the population does not become a safer one. At any moment of time, the LEARNING controller may become unstable and lead to the controlled system failure.

The experimental setup of the SAFE-LEARNING tandem is presented in Fig. 1. The numerical model described in section II simulates an inverted pendulum system. The SAFE controller is an LQG based controller. A linearized model of the IP dynamics was adopted for the purpose of designing the SAFE controller:

$$\begin{cases} (M+m)\ddot{p}(t) + m\frac{l}{2}\ddot{\Theta}(t) = C_V V(t) - (C_p + \beta)\dot{p}(t), \\ -m\frac{l}{2}\ddot{\Theta}(t) = m\dot{p}(t) \end{cases} \quad (1)$$

where *M* and *m* are rod and cart masses respectively, *l* is the rod length, *p(t)* is the cart position with respect to the center of the rail, $\Theta(t)$ is the rod angle with respect to the vertical, $C_V$ is the motor torque constant, $C_P$ and $\beta$ are the coefficients reflecting the dynamic and static friction in the coupling between the motor shaft and the rail. A detailed description of the SAFE controller can be found in [6].

The LEARNING controller is a multi-layer perceptron (neural network) composed of the neurons with the sigmoid transfer function. The neural controller has a fixed architecture:
a. Four inputs – cart position in meters (from –0.5 to 0.5 in respect to the center of the rail); cart velocity in meters per second (from -5.0 to 5.0); rod angle in radians (from -0.5 to 0.5 with respect to the vertical) and rod angular velocity in radians per second (from –5.0 to 5.0).
b. Two hidden layers, 4 and 2 neurons each.
c. One output. The neurons used in this neural network could only provide output in the range from 0.0 to 1.0, which is later scaled to the range from 0.0 to 5.0 (motor control voltage).

The weights of the neural controller were represented in the GA population as a binary string 1056 (33 weights x 32 bits) bits long. Each weight of the neural network was encoded by a 32-bit fixed-point number (24 bits for the integer part and 8 bits for the fractional part).

The switching block monitors the state of the controlled system and switches control to the SAFE controller as the system approaches a dangerous state. This block also controls the GA block by providing the signal for switching to the next specimen. In addition, it brings the system to some initial state, so each specimen of the GA population has approximately equal testing conditions. In this case, the switching block controls the SAFE controller, which in turn positions the cart in the middle of the rail and balances the rod. The SAFE controller has 5 seconds to return the IP system to the initial state.

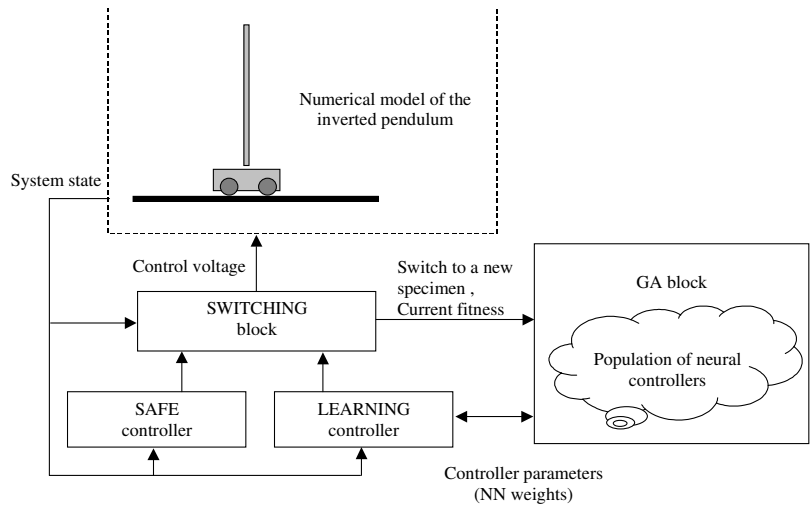

**Fig. 1:** SAFE/LEARNING tandem setup.

The switching block is probably the most important part of the suggested method. The stability of the controlled system depends upon timely switching from LEARNING to SAFE controller when the current state still allows the SAFE controller to recover. The system's performance is unacceptable if the switching is too late and the SAFE controller is not able to recover. At the same time, reducing the size of the dynamic region in which the LEARNING controller is allowed to operate is also undesirable. If the LEARNING controller does not have enough range in which to operate, the efficiency of the GA optimization may degrade beyond practical limits.

Fig. 2 illustrates a possible implementation of the switching algorithm. A N-dimensional hypercube (operating region of the LEARNING controller) is allocated within the asymptotic stability region of the SAFE controller. The hypercube is specified by the switching limits in respect to the given initial state of the system $S_0$. The switching limits are set in such manner that for the specified sampling time interval $\Delta T$ and the current state of the system $S(t)$ belonging to any boundary point of the hypercube the next state $S(t+\Delta T)$ will remain within the asymptotic stability region of the SAFE controller. The LEARNING controller is given control from the initial state $S_0$. During operation of the LEARNING controller the switching block checks the state of the system with the discrete time interval $\Delta T$. If the current state of the system is out of the specified limits, the control is turned over to the SAFE controller, which returns the system to the initial state $S_0$.

The SAFE/LEARNING tandem operates as follows:
1. The GA population is randomly initialized.
2. Evolutionary learning is performed for a given number of generations $G$:
   a. The switching block connects the SAFE controller to the IP system for 5 seconds to set the initial conditions.
   b. Weights of the LEARNING controller are set from the next specimen from the GA population.
   c. The switching block connects the LEARNING controller to the IP system. The LEARNING controller controls the inverted pendulum system for the next $T$ seconds.
   d. The switching block monitors the state of the IP system and switches to the SAFE controller if a dangerous state is detected or after $T$ seconds. It also computes the fitness value that characterizes current parameters of the LEARNING controller:
      i. If no dangerous states were detected during time period $T$, the fitness value $F$ for the current controller is calculated as:

$$F = \int_0^T \left(\frac{P(t)}{P_W}\right)^2 + \left(\frac{A(t)}{A_W}\right)^2 dt, \quad (2)$$

where $P(t)$ is cart position in meters, $P_W$ is position weight coefficient in meters, $A(t)$ is rod angle in radians, $A_W$ is angle weight coefficient in radians.

   ii. If the switching block detects a dangerous state at time $M$ and switches control to the SAFE controller, the fitness value for the current controller is equal to:

$$F = \int_0^M \left(\frac{P(t)}{P_W}\right)^2 + \left(\frac{A(t)}{A_W}\right)^2 dt + \\ (0.1T + T - M)\left(\frac{P_M}{P_W}\right)^2 + \left(\frac{A_M}{A_W}\right)^2, \quad (3)$$

where $P(t)$, $P_W$, $A(t)$, $A_W$ are the same as in 2.d.ii; $P_M$ is maximum possible cart displacement in meters, $A_M$ is maximum possible rod angle in radians.

The formulas (2) and (3) define the fitness (and performance) of a controller is such way that a controller with better RMS error of the cart position and rod angle will have a lower value of $F$.

e. Upon expiration of the $T$ seconds or a switch to SAFE, the fitness value is stored (replacing the previously stored value) by the GA block as an attribute of the current specimen. The current specimen number is checked against the population size. If the current specimen is not the last specimen in the population then the procedure repeats from step 2.a. If the current specimen is the last specimen in the population, the procedure goes to step 2.f.
f. At the new generation, the evolutionary manipulations are performed on the population:
   i. Reproduction by tournament selection. The 5% of the population with the highest fitness are carried through independently. Another 5% percent of the population is reinitialized randomly.
   ii. Crossover operation is performed with probability 0.5.
   iii. Mutation is performed with probability 0.1.
g. If the number of generations is less than a predefined limit *G*, the simulation continues from 2.a. If the number of generations is over the limit, the simulation proceeds to step 3.
3. Selection of the best specimen is performed for *S* generations:
   a. The switching block connects the SAFE controller to the IP system to set the initial conditions.
   b. Weights of the LEARNING controller are set from the next specimen from the GA population.
   c. The switching block connects the LEARNING controller to the IP system. The LEARNING controller controls the inverted pendulum system for the next *T* seconds.
   d. Switching and fitness computations are computed identically to step 2.d.
   e. Upon expiration of the *T* seconds or a switching event, the fitness value is accumulated (by addition to the previously stored value) by the GA block as an attribute of the current specimen. The current specimen number is checked against the population size. If the current specimen is not the last specimen in the population then the IP model is reset to its initial state and the procedure repeats from step 3.a. If the current specimen is the last specimen in the population, the procedure goes to step 3.f.
   f. If the number of generations is less that *G+S*, the simulation continues from 3.a. If the number of generations is over the limit, the simulation proceeds to step 4.
4. Accumulated fitness is sorted and the specimen with the best (lowest) fitness value is saved. The saved specimen represents the best controller that was produced during the simulation.

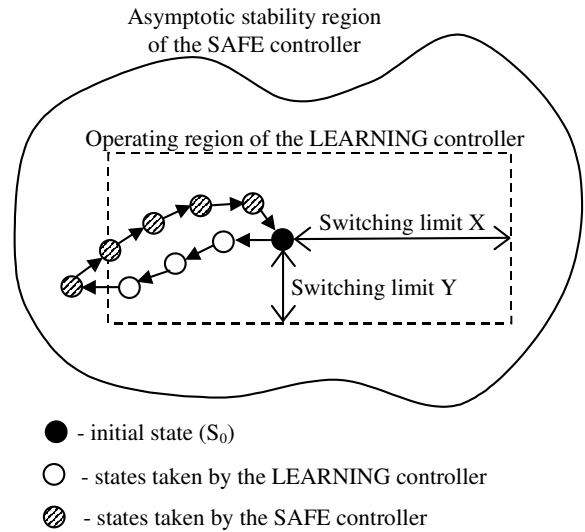

**Fig. 2**: Switching from the LEARNING to the SAFE controller (a 2-dimensional example)

The procedure above was normally performed for *G*=500 generations with the population size of 100 specimens. Specimen selection was performed for *S*=20 generations.

## IV. RESULTS

The experiments were conducted by setting the switching limits to ±10cm for cart position, ±20cm/sec for cart velocity, ±5.7 degrees for rod angle and ±115 degrees/sec for angular velocity. These switching values provided an operating region within of the asymptotic stability region of the SAFE controller that allowed enough space for operation of the GA optimizer. The GA optimization of neural controllers with the described parameters always produced bang-bang controllers. Fig. 3 and Fig. 4 illustrate the difference between balancing by the SAFE and an optimized LEARNING controller. Note that the neural controller develops higher cart velocities and rod angular velocities during balancing. Cart position, rod angle and motor control voltage acquired from a neural controller optimized by the genetic algorithms are shown in Fig. 5. The IP system did not fail (reach the end of the rail or drop the rod) during 50 test runs.

The average parameters of the produced neural controllers compared to the SAFE controller are given in Table I. The RMS values for the neural controllers were measured during 30-second rod balancing and compared to the RMS values of the SAFE controller (cart position RMS=0.01391m, rod angle RMS= 0.7133 degree).

The optimized neural controllers on average offered approximately 21% lower RMS of the rod angle and 49% lower RMS of the cart position than the SAFE controller. The best specimen produced by the GA optimization had 2.5 times lower RMS of cart position (59% reduction) combined with 22% reduction in rod angle RMS. The weight coefficients allowed placing the emphasis either on optimization of the cart position or rod angle.

## V. CONCLUSIONS

The optimization method, described in this paper, was successfully applied to failure-free controller optimization by genetic algorithms. A clear advantage of the suggested method is that it can operate directly on the controlled system, thus accounting for the existing non-linearities and uncertainties of the parameters without jeopardizing the safety. The unacceptable solutions in the parameters of the LEARNING controller that appear in the course of genetic optimization do not lead to the system's failure due to the presence of the SAFE controller, which recovers the system from the dangerous states. Conducted experiments have shown that an optimized neural controller can be produced by genetic algorithms on an inherently unstable system (inverted pendulum) without failures of the IP system.

Application of the suggested method is not limited only to optimization of the neural controllers. The method can be applied almost to any type of a system controller (neural, fuzzy, classic, etc.) that can benefit from optimization directly on the system.

Further research is necessary to determine proper functioning of the switching block (related to the asymptotic stability region of the SAFE controller) in such a way, that the SAFE controller is guaranteed to recover from any dangerous situation while providing the LEARNING controller the maximum available space for learning its behavior.

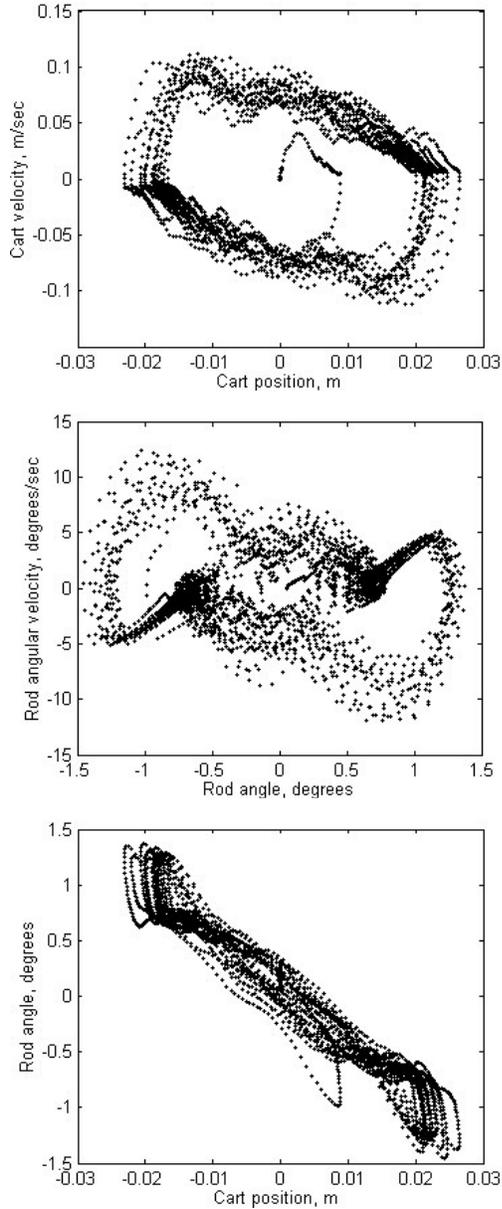

**Fig. 3:** Two-dimensional projections of system variables during 30-second balancing by the classic controller

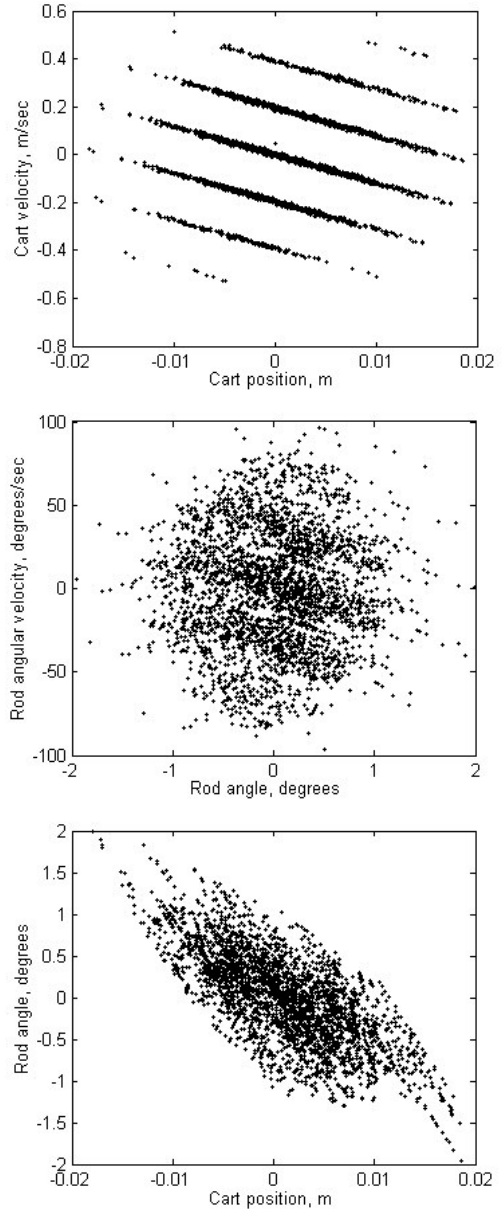

**Fig. 4:** Two-dimensional projections of system variables during 30-second balancing by a neural controller

The simplified approach of allocating an N-dimensional hypercube (as applied in this paper) could be replaced by a more advanced method, such as building a predictor that maps the stability region of the SAFE controller during its independent operation and decides whether the current state is a dangerous one. Also it should be noted that the LEARNING controller could not potentially cover the whole state space of the system. Rather, it learns its behavior on a subspace defined by the switching parameters. Thus, the switching limits should either include all the states that an optimized controller will encounter during its operation or additional measures (such as switching back to the SAFE controller) should be taken during its operation.

## VI. ACKNOWLEDGEMENTS

The authors wish to acknowledge the support provided for this work by Allegheny Power.

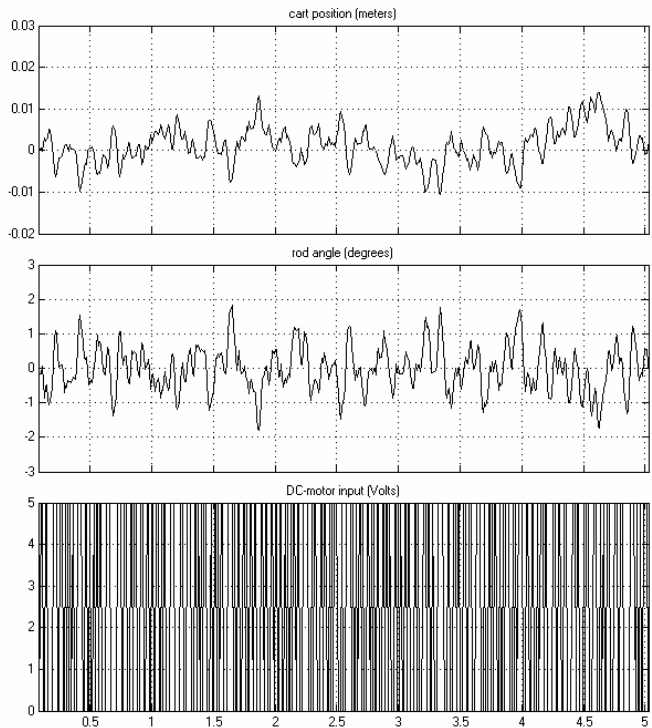

**Fig. 5:** Cart position, rod angle and motor control voltage acquired from an optimized neural controller during a 5-second balancing interval

**Table 1:** The average RMS of cart position and rod angle obtained on the neural controllers optimized with different weight coefficients $P_W$ and $A_W$

| $P_W$, cm | $A_W$, degrees | Cart position RMS, cm | Rod angle RMS, degrees | Reduction in cart position RMS, % | Reduction in rod angle RMS, % |
|---|---|---|---|---|---|
| 0.5 | 2.0 | 0.6527 | 0.6431 | 53.07 | 9.83 |
| 0.5 | 1.0 | 0.6553 | 0.6131 | 52.89 | 14.04 |
| 0.5 | 0.5 | 0.692 | 0.5705 | 50.25 | 20.01 |
| 1.0 | 0.5 | 0.7759 | 0.5048 | 44.22 | 29.22 |
| 2.0 | 0.5 | 0.8508 | 0.4795 | 38.83 | 32.76 |